\documentclass[a4paper,twocolumn,10pt]{article}
\usepackage{eurosis}
\usepackage{url}
\usepackage{natbib}
\usepackage{graphicx}
\usepackage{booktabs}
\usepackage[justification=centering,width=0.8\textwidth]{caption}
\usepackage{hyperref}
\usepackage{newtxtext,newtxmath}

\bibpunct[; ]{(}{)}{,}{a}{}{;}

\begin{document}

\title{UNDERSTANDING PLAYER ENGAGEMENT AND IN-GAME PURCHASING BEHAVIOR WITH ENSEMBLE LEARNING}

\author{Anna Guitart, Ana Fern\'andez del R\'io and \'{A}frica Peri\'a\~{n}ez\\
Yokozuna Data, a Keywords Studio\\
Chiyoda 5F, Aoba No. 1 Bldg., Tokyo, \\
Japan\\
E-mail: \{aguitart, afdelrio, aperianez\}@yokozunadata.com}

\date{}

\maketitle

\thispagestyle{empty}


\begin{abstract}
As video games attract more and more players, the major challenge for game studios is to retain them. We present a deep behavioral analysis of churn (game abandonment) and what we called ``purchase churn'' (the transition from paying to non-paying user). A series of churning behavior profiles are identified, which allows a classification of churners in terms of whether they eventually return to the game (false churners)---or start purchasing again (false purchase churners)---and their subsequent behavior. The impact of excluding some or all of these churners from the training sample is then explored in several churn and purchase churn prediction models. 
Our results suggest that discarding certain combinations of ``zombies'' (players whose activity is extremely sporadic) and false churners has a significant positive impact in all models considered.
\end{abstract}

\section{INTRODUCTION}
\label{intro}

The concept of \emph{churn} is as old as the customer--service relationships themselves. Churn occurs when a certain user stops using a service, i.e.\ when the relationship between the customer and the service provider ends \citep{mozer2000churn}. This term is widely used in a variety of industries including retail banking \citep{mutanen2006customer}, telecommunications \citep{hwang2004ltv} and gaming \citep{runge2014,perianez2016churn}.

Churn remains one of the most important metrics to evaluate a business, as it is directly linked to user loyalty \citep{hwang2004ltv}. High retention (i.e.\ low churn) points to a healthy business, and increases in user retention usually translate into higher revenues. In free-to-play games retention is crucial, since many of them have in-app purchases as their main source of revenue and, moreover, gaining new users through marketing and promotion campaigns is typically much costlier than retaining existing players \citep{Monetization}.

If there is a contractual relationship with the customer (as is normally the case in sectors such as telecommunications, see \citealt{mozer2000churn}), the definition of churn is unambiguous: it happens when the customer cancels the contract or unsubscribes from the service. On the other hand, when there is no contract (or equivalent relationship) it is more difficult to assess whether a user has really churned or not.
The appropriate way of defining churn in this kind of commercial activities must be carefully studied in light of their particularities and needs, and also of the purpose of the definition itself. This is the situation that applies to online games \citep{perianez2016churn,GameBigData,cig2018competition,chen2019competition}, where most users stop playing without deleting their account. Additionally, free-to-play gamers who are active but make no purchases are of little or no economic value, and this allows us to introduce another type of churn definition within the video game context: {\it purchase churn}, which refers to paying users who cease to spend money on the title and is as tricky to define as conventional churn. (We will occasionally refer to the latter as \emph{login churn}, for clarity.)

The usual strategy is to consider that a player has churned after a certain number of days of inactivity~\citep{runge2014,perianez2016churn}. Here we begin by examining how to choose a suitable (login/purchase) churn definition (in terms of days without activity/purchases). The goal of classifying players into active or churned is twofold: On the one hand, to have an accurate measure of the current health of the game. On the other, to label players in an appropriate way to successfully train churn prediction models. 

Accurately predicting churn is of paramount importance for any business. In video games, the early detection of potential (login or purchase) churners may give studios the chance to target players individually---with personalized discounts, presents or contents---in an attempt to re-engage them. Previous works addressing churn prediction in video games have treated churn either as a classification \citep{sifa2015predicting,chen2019competition} or survival problem \citep{perianez2016churn,chen2019competition}, with the latter approach being especially well-suited due to the censored nature of churn. Other related works used churn predictions to compute the lifetime value of individual players \citep{chen2018ltv}.

Player profiling (i.e.\ grouping users based on their behavior) is another noteworthy problem \citep{bauckhage2014clustering,drachen2012guns,drachen2014comparison,saas2016discovering}, which we also address here from a churn perspective. Our main goal is to characterize players who are identified as churners but eventually start playing again, namely {\it false churners}. Some of them are \emph{genuine false churners}: in spite of meeting the corresponding churn definition, they never left the game, but just remained inactive for a relatively long time. Others (those who had a lengthier period of inactivity before returning to the game and thus can be considered to actually have churned) are more rightly regarded as \emph{resurrected} players. 
In contrast, we will regard \emph{all} players who start purchasing again after a prolonged lapse without spending any money as \emph{purchase resurrected}.
There is yet another group of interest in connection with churn: players whose activity is so sporadic that---regardless of whether or not they have been tagged as churners in the past based on the particular churn definition used---they can hardly be deemed as active users; we will refer to them as \emph{zombies}. Such a classification of players according to their churn behavior is interesting on many levels, but in this work we focus on assessing its impact on the accuracy of churn prediction models.

The remainder of the paper is organized as follows: First we introduce the two main standard approaches used to define churn in video games, as well as the specific dataset and definitions adopted in our experiments. Then, we describe the churn prediction models analyzed in this study. Finally, after presenting and discussing the prediction results obtained by discarding different types of churners, we provide a brief summary of our findings and deliver our conclusions.

\subsection{Our Contribution}
To the best of our knowledge, this is the first work to simultaneously address login and purchase churn prediction, compare the classification and survival approaches and study the effect of excluding different kinds of churners from the training on the accuracy of the results.

\begin{figure*}[ht!]
  \centering
  \includegraphics[width=0.45\textwidth]{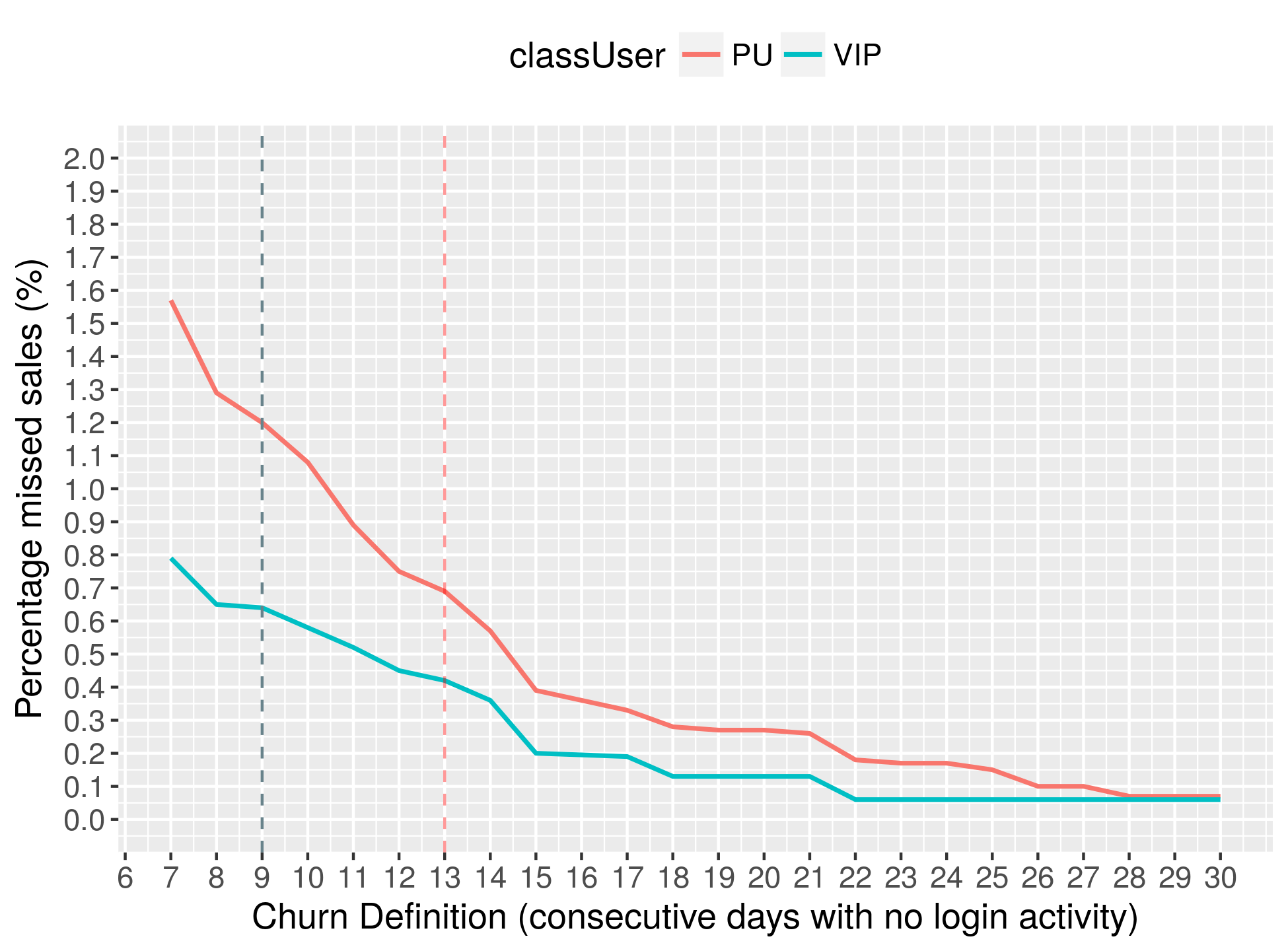} 
  \includegraphics[width=0.45\textwidth]{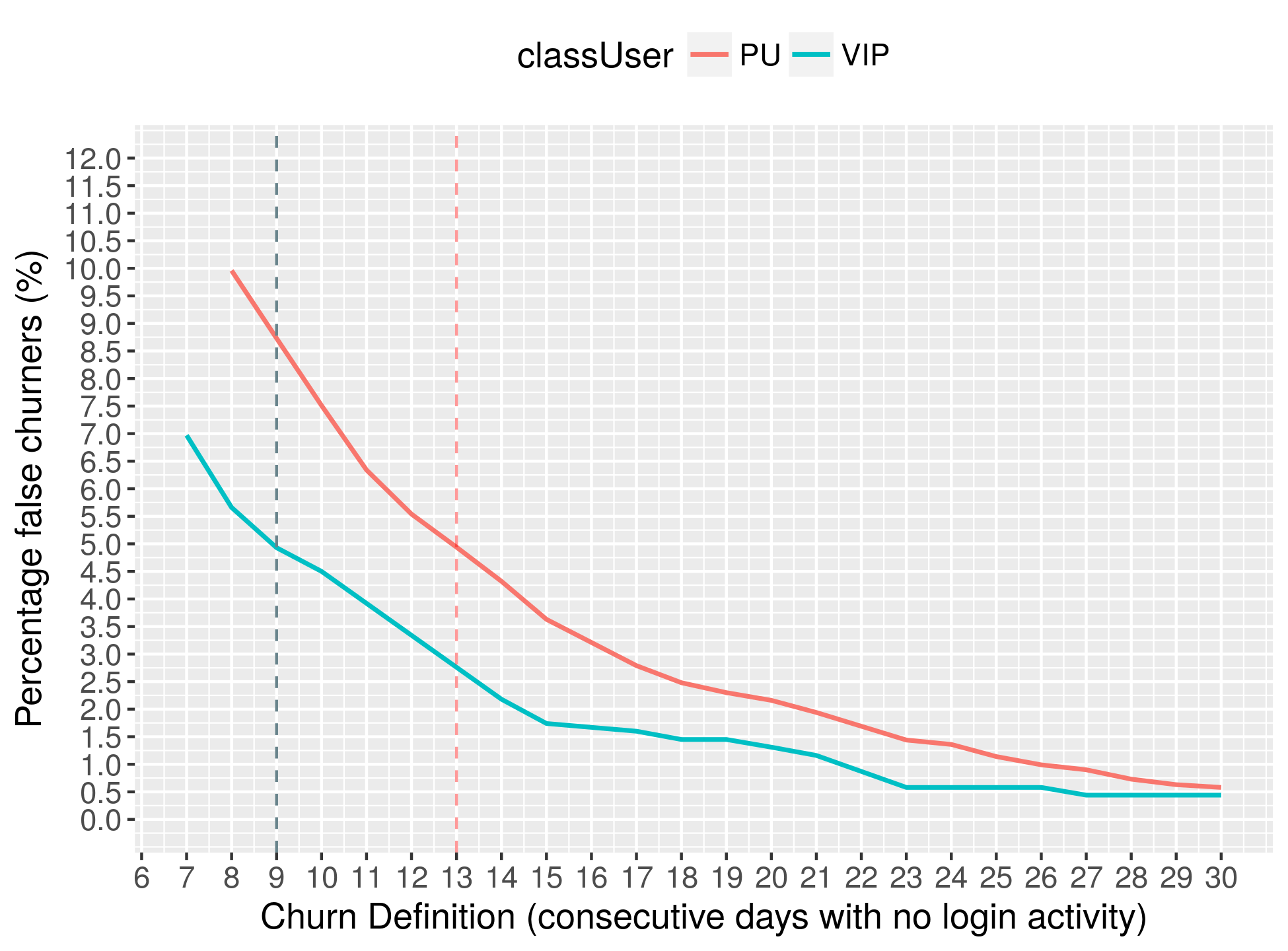} 
\caption{Determination of the login churn definition for VIP players (blue) and all paying users (PUs, red) based on two indicators: the percentages of missed sales (left) and false churners (right) during the first two months of data. By imposing these percentages to remain below 1\% and 5\%, respectively, we obtain 9 days (for VIP players) and 13 days (for PUs) as the inactivity period after which a player is considered to have churned.}
\label{churnDef}
\end{figure*}

\section{DEFINITIONS AND DATASET}
\label{defAndDataset}

\subsection{Defining Churn}
\label{defs}

Two main approaches to define churn in terms of player inactivity can be found in the literature:

1) Using a \emph{fixed time window} for all players \citep{cig2018competition}. For example, we could consider players who logged in during the previous month but not within the current one to be churners. This kind of strategy can be useful for some purposes---such as tracking game retention over long time scales---but it is not without shortcomings. In particular, it is fairly insensitive to specific player connection patterns, something especially problematic for churn prediction. 

2) Using a \emph{moving time window} for each player. To overcome the limitations of the above approach, most works measure the churn-defining inactivity period through a moving window, i.e., referred to individual player time instead of calendar time~\citep{runge2014,perianez2016churn}. While this method is computationally more demanding, it is also much better suited to model churn risk, and is thus the one followed in this paper.

The length of the optimal moving time window is highly game dependent. While in very casual titles a few days of inactivity typically signal a real user disengagement, in massively multiplayer online role-playing games time between sessions is usually much longer, and so longer time windows are required to correctly identify churners. The situation is analogous for purchase churn, as the typical purchase frequency may also vary greatly from game to game.

In this work, window lengths are selected so as to minimize two quantities: the \emph{percentage of false churners} (number of churners who eventually return to the game over total number of churners) and the \emph{percentage of missed sales} (sales from false churners after they return to the game over total sales). Considering long enough time windows can make both of these quantities vanish. However, our aim is to detect churn as soon as possible, both to have an accurate picture of player engagement at any given time and to have sufficient room for manoeuvre to try and re-engage potential churners. In particular, for our churn definitions we consider the shortest period of inactivity that keeps false churners under 5\% and missed sales under 1\% (although these figures can be fine-tuned according to the specific requirements of the analysis). Further details are given below.

\subsection{Dataset}
\label{dataset}

We used game data from the Japanese title \emph{Age of Ishtaria} (a free-to-play, role-playing mobile card game developed by Silicon Studio), collected between 2014-10-02 and 2017-05-01. The data contains detailed daily information about each player, including level-ups, playtime, purchases and sessions. 

Only top spenders (\emph{VIP players} or \emph{whales}) were considered, as they are the most valuable users. 
We define VIP players as those with total outlay above a certain threshold (computed from the first two months of data so that whales provide at least 50\% of the total revenue) and there were around 6000 of them in the studied dataset. 

Data from other mobile games were also evaluated following the same methodology, and we obtained equivalent results, which shows the applicability of the proposed concepts to online games. These results are not included in the paper due to space limitations. 
In the case of non-online games, similar principles could be applied. However, as the purpose of this work is to give a solution that can be used in an operational environment, we focused on studying online games, where actions can be actively performed on the players and player information is continuously updated.

\begin{figure*}[ht!]
  \centering
  \includegraphics[width=0.3\textwidth]{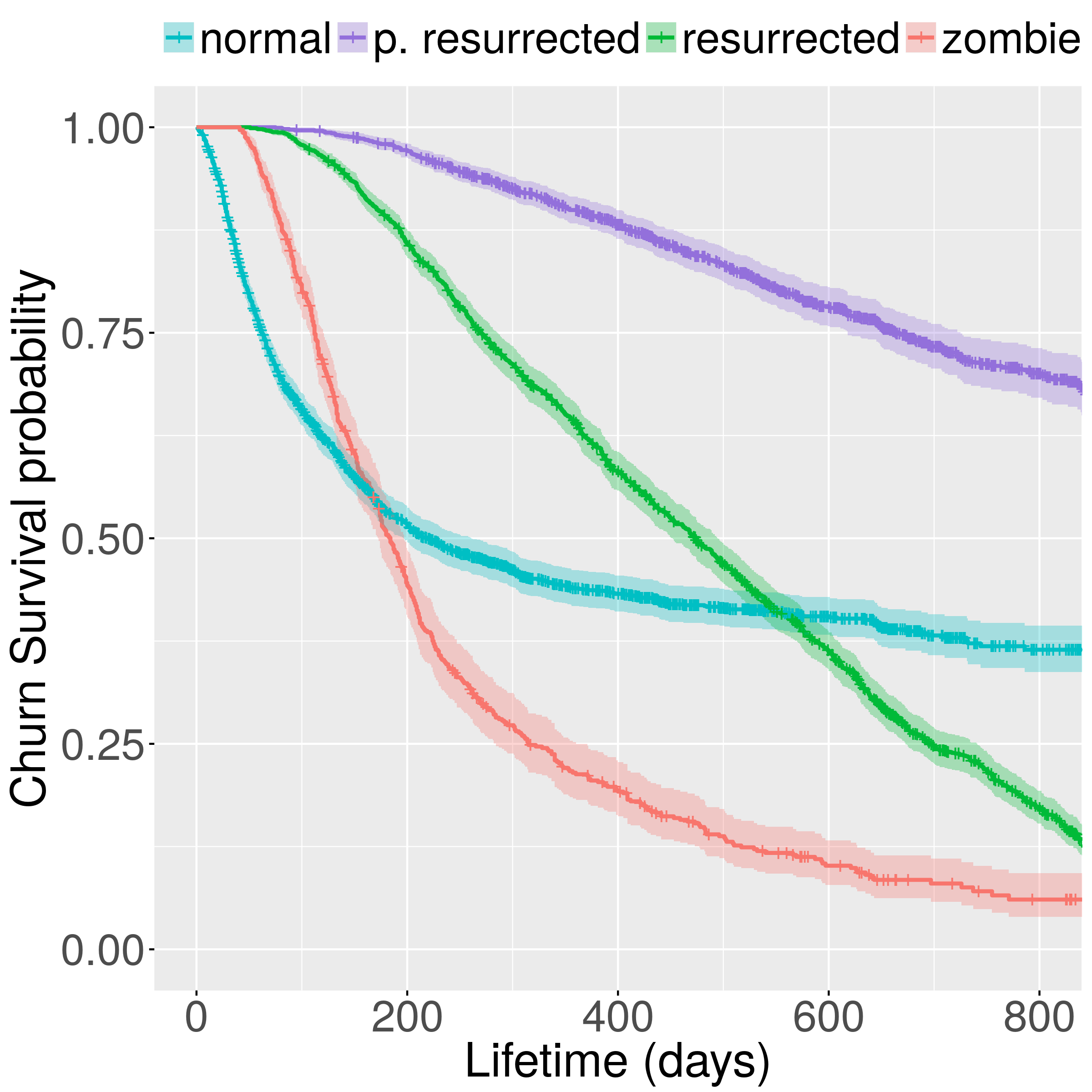} 
  \includegraphics[width=0.3\textwidth]{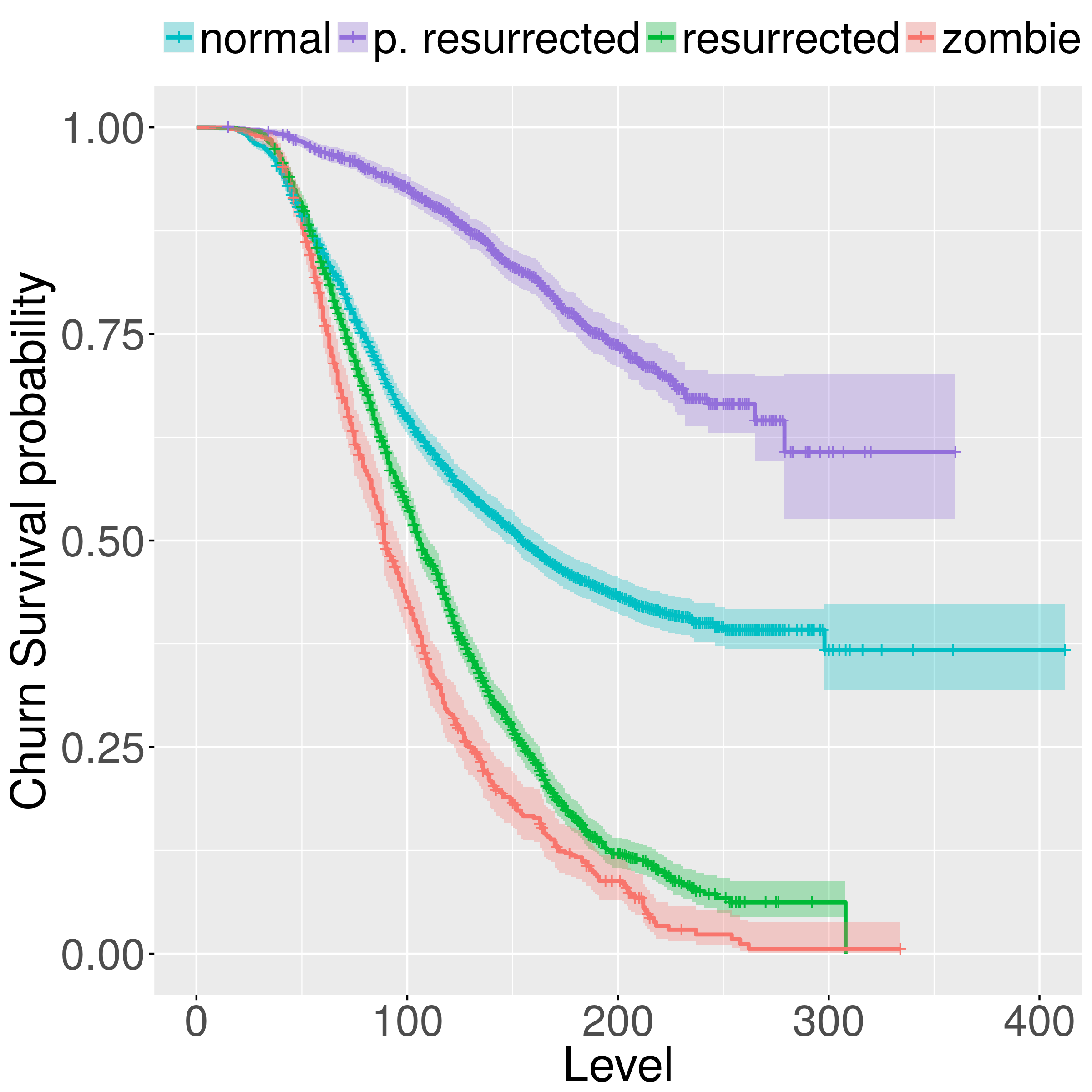} 
  \includegraphics[width=0.3\textwidth]{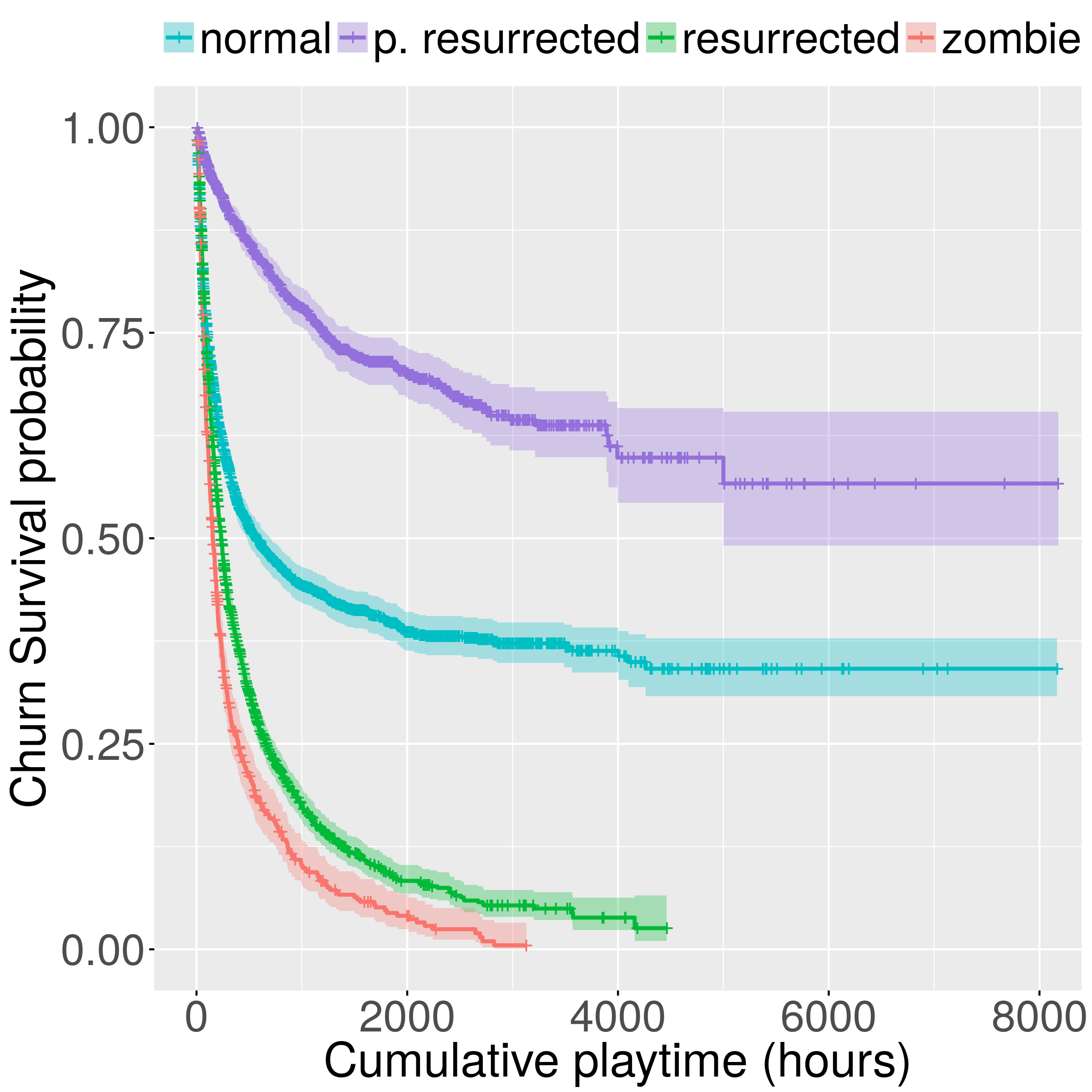} \\
  \includegraphics[width=0.3\textwidth]{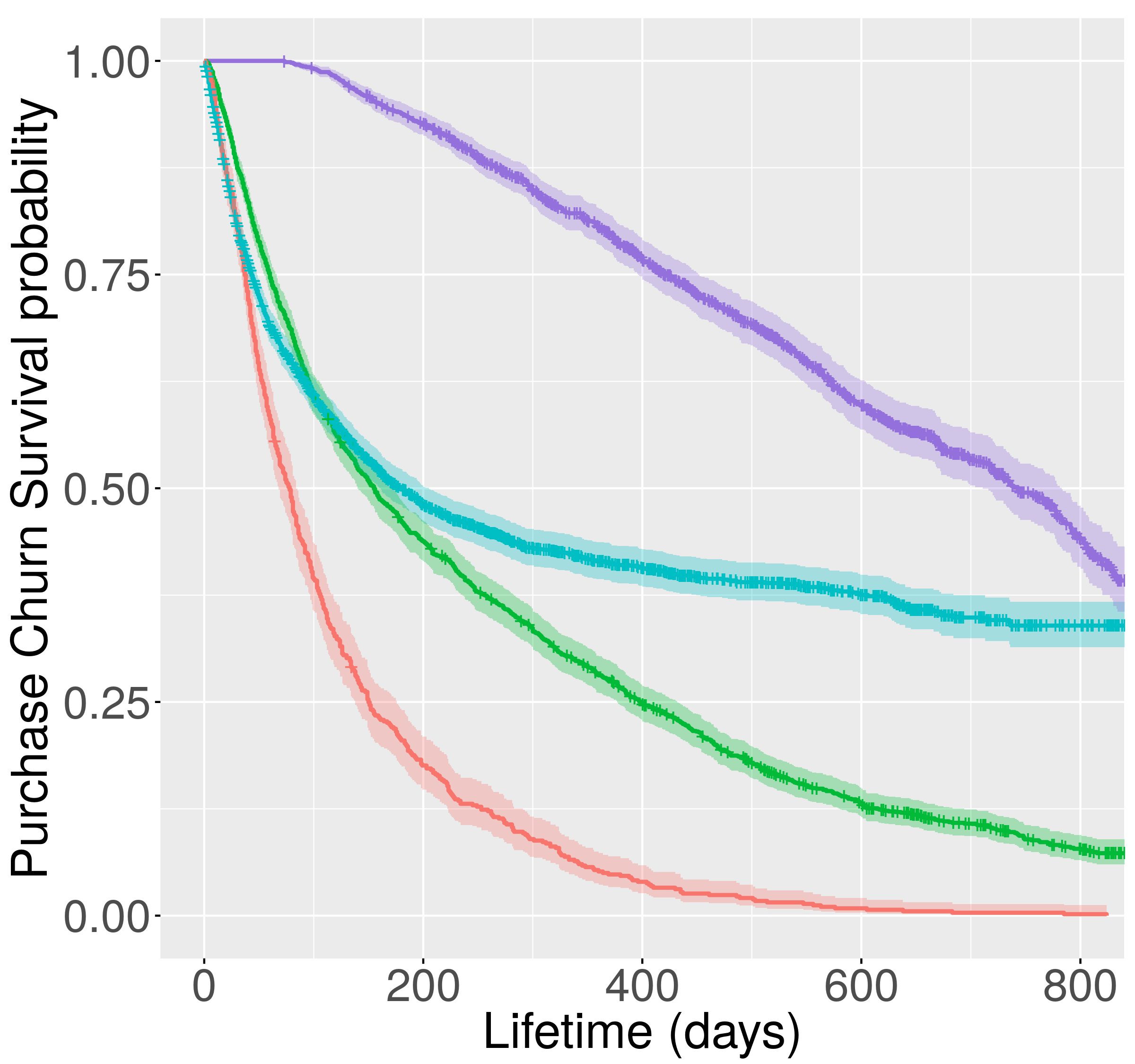} 
  \includegraphics[width=0.3\textwidth]{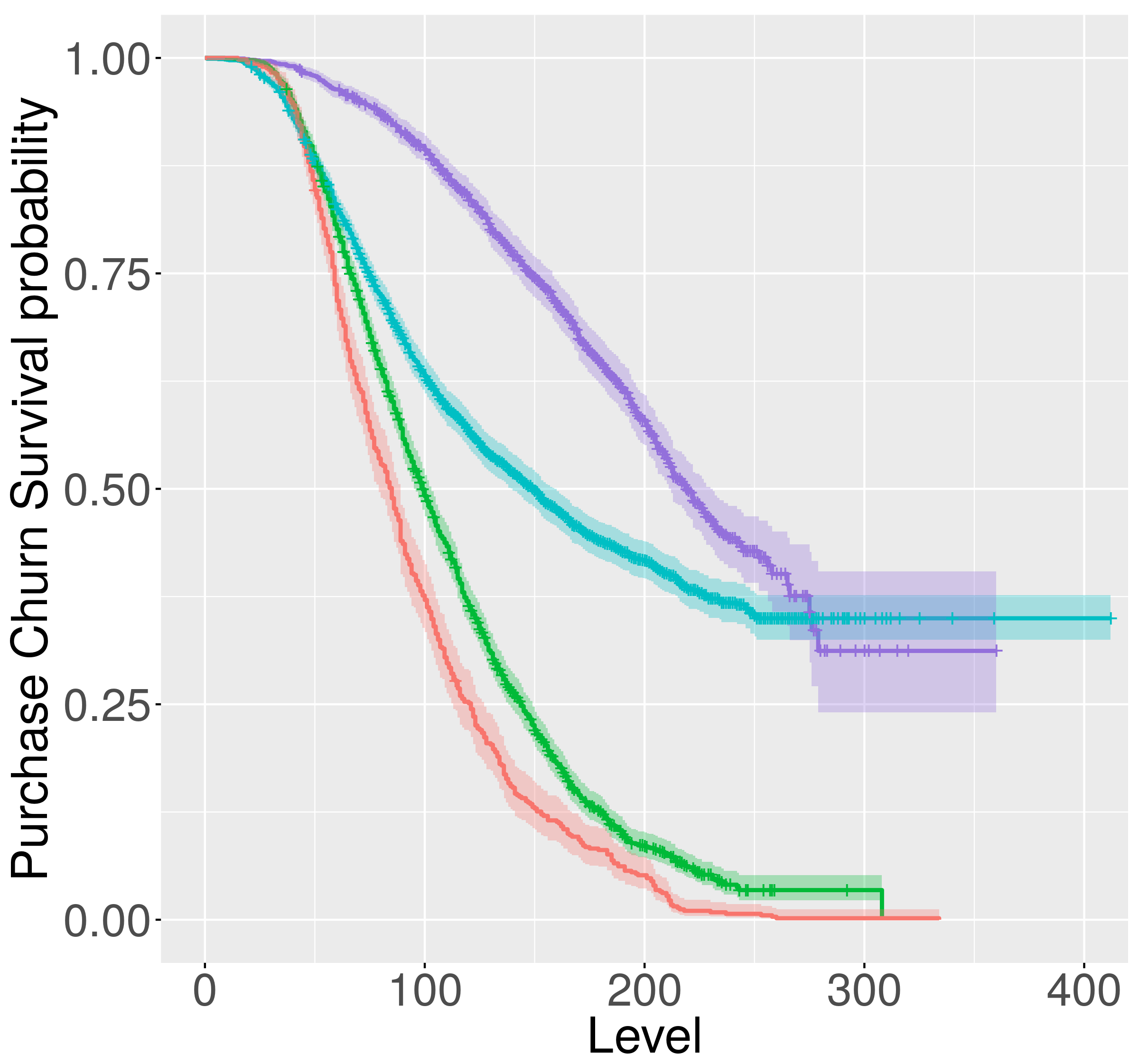} 
  \includegraphics[width=0.3\textwidth]{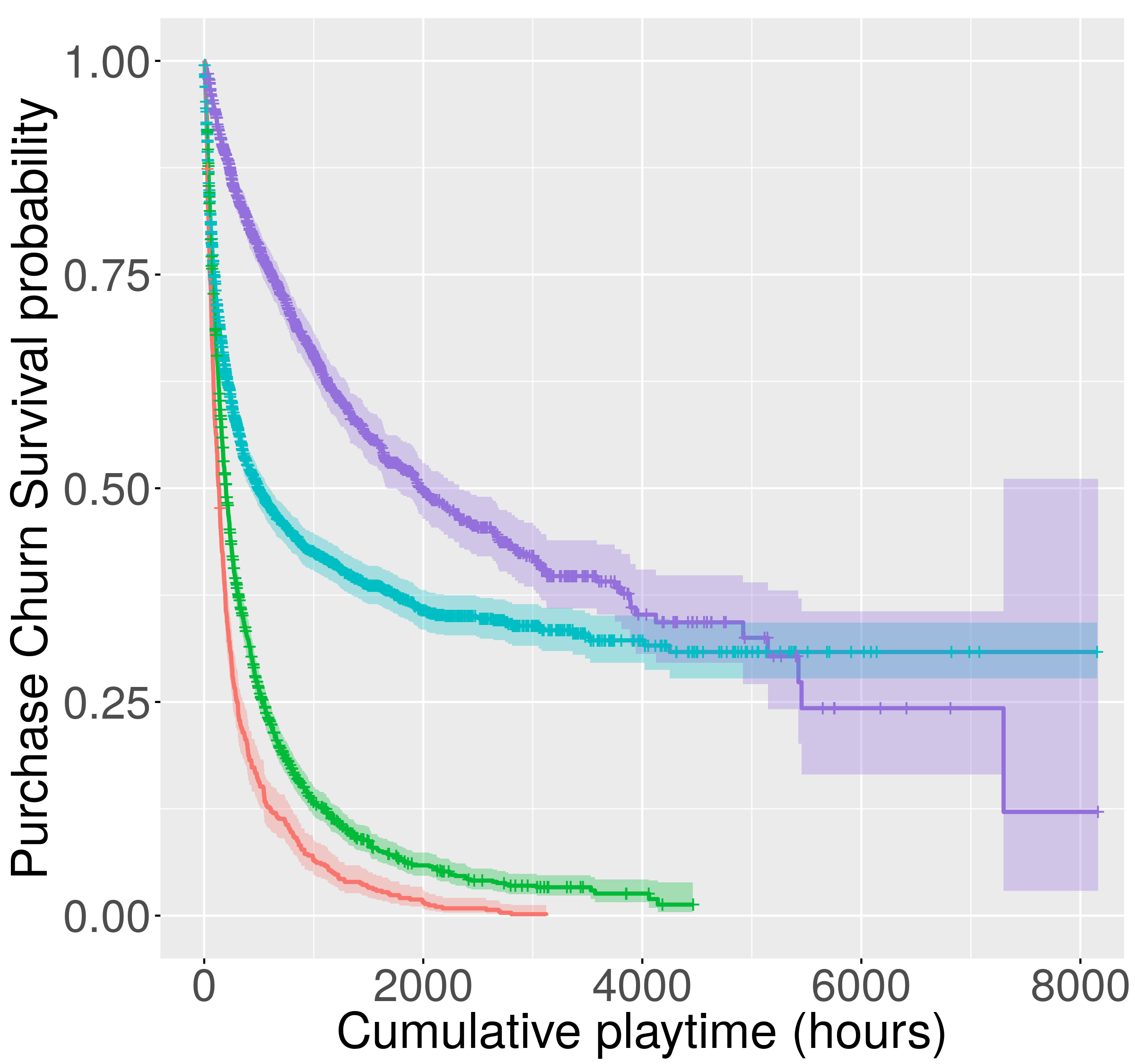} \\
\caption{Cumulative survival probability (Kaplan--Meier estimates) as a function of time since first login (left), game level (center) and cumulative playtime (right) for VIP players. Top/bottom panels refer to login/purchase churn. Curves are stratified by churner type: \emph{normal}, \emph{zombie}, \emph{resurrected} and \emph{purchase resurrected} players. Shaded areas represent 95\% confidence intervals. }
\label{kaplanMeier_playerType}
\end{figure*}

\subsection{\emph{Age of Ishtaria}'s Churn Definition}
\label{churndefcomputation}

Figure \ref{churnDef} shows graphically how the \emph{login churn} definition was inferred from the first two months of data. The percentage of missed sales (left) and percentage of false churners (right) were evaluated for different churn definitions---letting the inactivity period after which a player is considered to have churned vary between 3 and 90 days---when considering all paying users (PUs, red curves) or just VIP players (blue curves). As already discussed, we require these percentages to be less than 1\% and 5\%, respectively, which yields an inactivity period of 13 days for all PUs and 9 days for VIP players only. Since our analysis is restricted to top spenders, we will use the latter time window as our churn definition. Note that the percentage of false churners will tend to increase when considering extended data periods (longer than two months) but, for practical reasons, it is desirable to set the churn definition as soon as possible. 
In any case, we checked that such increase was not very significant---the percentage remained well below 10\% even for longer windows of 6 months, taken at different dates across the full dataset---which means that the two-month data are representative of the overall churning behavior and supports our strategy. (Note that our real aim here is to restrict the number of \emph{genuine} false churners. Thus the percentage of false churners can be higher when considering the whole dataset, due to the increase in the number of resurrected players.) 

Following a similar approach we found that \emph{purchase churn} should be defined as 50 days without any spending for VIP users. This inactivity period is much longer than in the previous (login) case, and thus a much larger (roughly by a factor 5) sample is needed to properly determine it. In practice, for a new title, it is possible to obtain a first working definition by using other games as reference, and then revisit it when a large enough data sample is available. 

\subsection{Churner Profiling}
\label{typesOfPlayers}
Three different groups of players with a particularly interesting churn-related behavior will be considered, and the impact of excluding them from the model training examined. These are

1) \emph{Resurrected} players: Those who return to the game after churning and remaining inactive for a prolonged period of time. When churn is defined as less than 10 days of inactivity (as in our case), we require that period to be of at least 30 days. 
Users who return to the game before 30 days of inactivity are considered to be genuine false churners (i.e., to have been mistakenly marked as churners) rather than resurrected players.

2) \emph{Purchase resurrected} players:
In this study we identify all false purchase churners as purchase resurrected once they start spending again. (We thus disregard genuine false purchase churners.)

3) \emph{Zombies}: 
Players who exhibit a too disengaged behavior to be considered active users (but who are not churners at that moment). In this study, players with less than 3 hours of playtime, no level-ups and no purchases in the previous 30 days were labeled as zombies.

Players who do not fall into any of the previous three groups will be referred to as \emph{normal}. 

In the sample considered, 21\% of the players had churned and 5\% had purchase churned by the end of the data period. Around 10\% of all players were labeled as zombies, nearly 30\% as resurrected and 23\% as purchase resurrected at some point throughout their lifetime.
Although the high percentage of resurrected players could suggest that our churn definition was not restrictive enough, we should recall that its aim is to limit the presence of \emph{genuine} false churners rather than resurrected players (who typically churn for good shortly after returning to the game and thus do not increase the percentage of false churners in the long run). 

Figure \ref{kaplanMeier_playerType} shows Kaplan--Meier survival curves for VIP players---as a function of playtime, lifetime (time since first login) and game level---stratified by user type. Purchase resurrected players have the highest survival probabilities against both churn and purchase churn. (The only exception could be purchase survival for very high game levels or playtime, where normal players seemingly have higher probabilities, although it is not possible to ascertain that due to the large uncertainties.) On the other hand, zombies have the lowest survival and purchase survival probabilities across all variables. Interestingly, for small lifetime (though not level or playtime) values, resurrected players present higher survival rates (against login churn) than normal players. After more than a year the trend is inverted, as the survival probability for normal players stabilizes whereas that of resurrected players continues decreasing at the same pace.
Note also that, in general, purchase survival curves are steeper than the corresponding (login) survival curves. This highlights the fact that all churners are also purchase churners (while the opposite is not true).

\section{MODELING}
\label{modeling}

We analyzed both binary and survival churn prediction models, exploring the effect of removing zombies, resurrected players, purchase resurrected players and combinations of them from the model training. The aim is to elucidate whether the presence of these players might be introducing noise that prevents the models from learning the {\it typical} VIP churn behavior more efficiently. 

To get the results shown in this paper, data until 2018-03-01 was used for training and the remaining data until 2018-05-01, for validation. Nonetheless, we also evaluated the impact of varying the training and validation data ranges, obtaining similar results in all cases.

\begin{table*}
	\centering
	\caption{Login and purchase churn prediction results for the binary and survival models, measured through the area under the curve (AUC) and the integrated Brier score (IBS), respectively. Survival results are given in terms of different predictor variables: lifetime, level and cumulative playtime. We consider different situations with regard to the training sample: including all users (\emph{none}) vs.~excluding zombie, resurrected or purchase resurrected players (or combinations of them). The best results for each model and variable are highlighted in bold.}
\small
\begin{tabular}{@{}ccccccccc@{}}
\toprule
    CHURN      & \multicolumn{2}{c}{Binary models (AUC)} & \multicolumn{6}{c}{Survival models (IBS)} \\
\cmidrule(lr){2-3}\cmidrule(l){4-9}    
excluding from training & by login & by purchase & \multicolumn{3}{c}{by login} & \multicolumn{3}{c}{by purchase}  \\  
\cmidrule(lr){4-6} \cmidrule(lr){7-9}
 & & & lifetime & level & playtime & lifetime & level & playtime \\ \midrule
none                               & 0.95 & 0.69 & 0.072 & 0.069 & 0.060 & 0.070 & 0.080 & 0.077\\
zombie                             & 0.93 & 0.69 & 0.034 & 0.047 & \textbf{0.035} & 0.055 & 0.067 & 0.086\\
resurrected                        & 0.90 & 0.68 & 0.043 & 0.048 & 0.041 & 0.070 & 0.080 & 0.080\\
p.~resurrected                      & 0.95 & 0.72 & 0.104 & 0.084 & 0.060 & 0.065 & 0.076 & 0.062\\
zombie, resurrected                & 0.94 & 0.69 & \textbf{0.029} & \textbf{0.041} & \textbf{0.035} & 0.055 & \textbf{0.057} & 0.086\\
zombie, p.~resurrected              & 0.93 & 0.72 & 0.057 & 0.068 & 0.049 & \textbf{0.053} & 0.067 & \textbf{0.050}\\
resurrected, p.~resurrected         & 0.92 & 0.73 & 0.071 & 0.068 & 0.057 & 0.065 & 0.068 & 0.057\\
zombie, resurrected, p.resurrected & 0.94 & 0.73 & 0.053 & 0.059 & 0.050 & \textbf{0.053} & \textbf{0.056} & \textbf{0.051}\\
\bottomrule
\end{tabular}
\label{resultsTable}
\end{table*}

\subsection{Model Specification}

Specifically, we investigated the performance of a \emph{conditional inference survival ensemble} model \citep{Hothorn06unbiasedrecursive}, described in detail in previous churn prediction studies \citep{perianez2016churn,GameBigData} of which the present work constitutes an extension. Player survival was described in terms of three different variables: playtime, lifetime (time since first login) and game level reached. On the other hand, binary classification was explored through \emph{conditional inference trees} \citep{Hothorn06unbiasedrecursive}. Ensembles of size 1000 were used in all cases. 

\subsection{Feature Selection}

Feature selection was also based on previous studies \citep{perianez2016churn,GameBigData} that constructed game-independent features measurable in most titles, such as playtime, purchases or number of actions of each player. 
We evaluated the best feature combination as a function of the model (binary or survival) and survival variable (lifetime, level, playtime). The possibility of adding a flag to identify the type of user (e.g.\ 1 for zombies and 0 for normal players) was also investigated. 
However, these variables proved to bias the models towards the behavior of the special users (affecting the accuracy of the predictions for normal players) and were discarded in the end. 

\subsection{Model Validation}

For conditional inference ensembles, model validation was performed through specific survival analysis error curves and the integrated Brier score (IBS) \citep{graf1999assessment}, in the way described by \cite{perianez2016churn}. The binary models performance was assessed using the area under the receiver operating characteristic curve (AUC); see e.g.~\cite{bradley1997}.

The set of players used for validation was the same in all cases (excluding zombies, resurrected and purchase resurrected players) so that we can fully assess the impact that training on different groups of users has on the predictions for the same group of players. This strategy was adopted to avoid massaging the data, which may lead to biased results.

\section{RESULTS}
\label{results}

\begin{figure*}[ht!]
  \centering
\hfill
\includegraphics[width=0.44\textwidth]{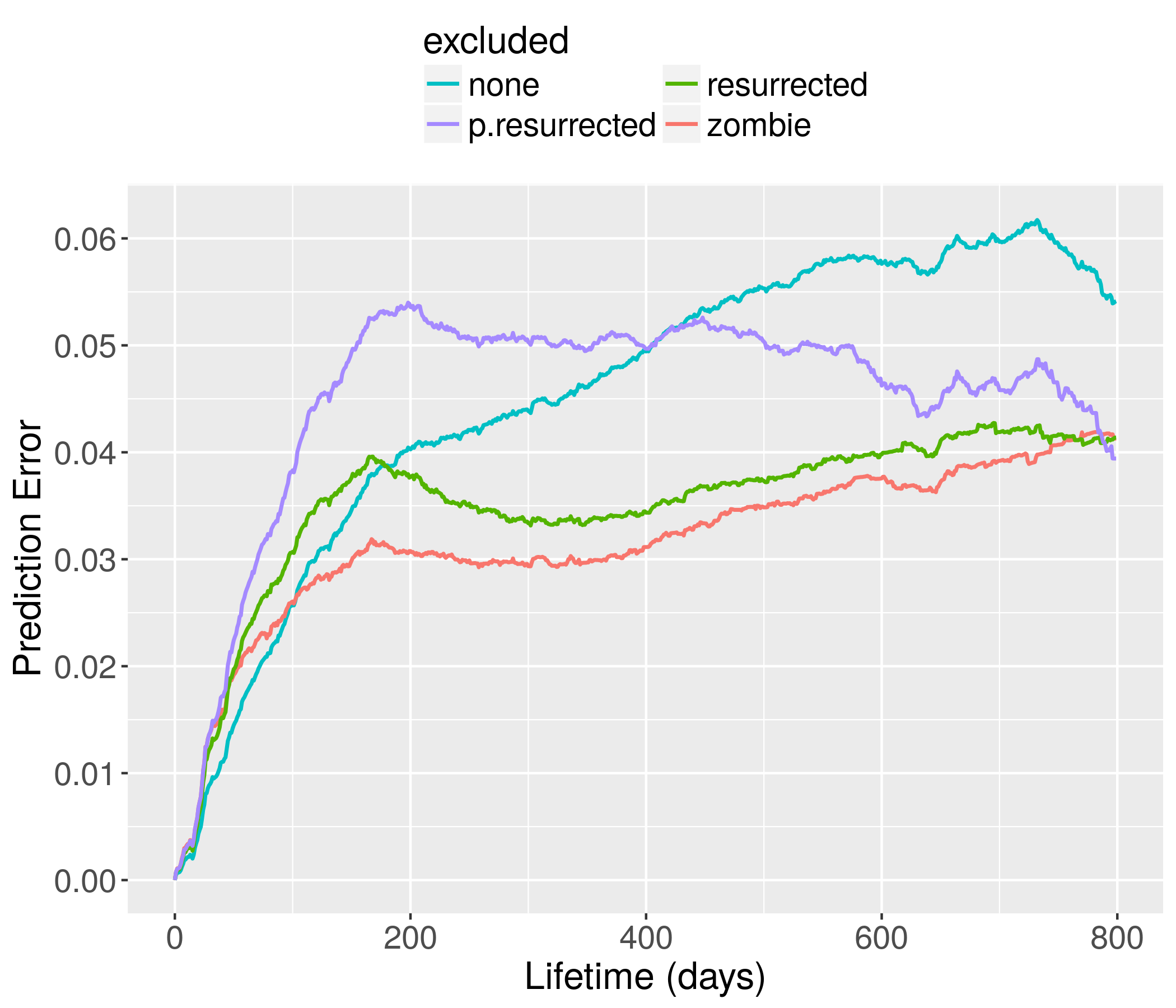}\hfill
\includegraphics[width=0.44\textwidth]{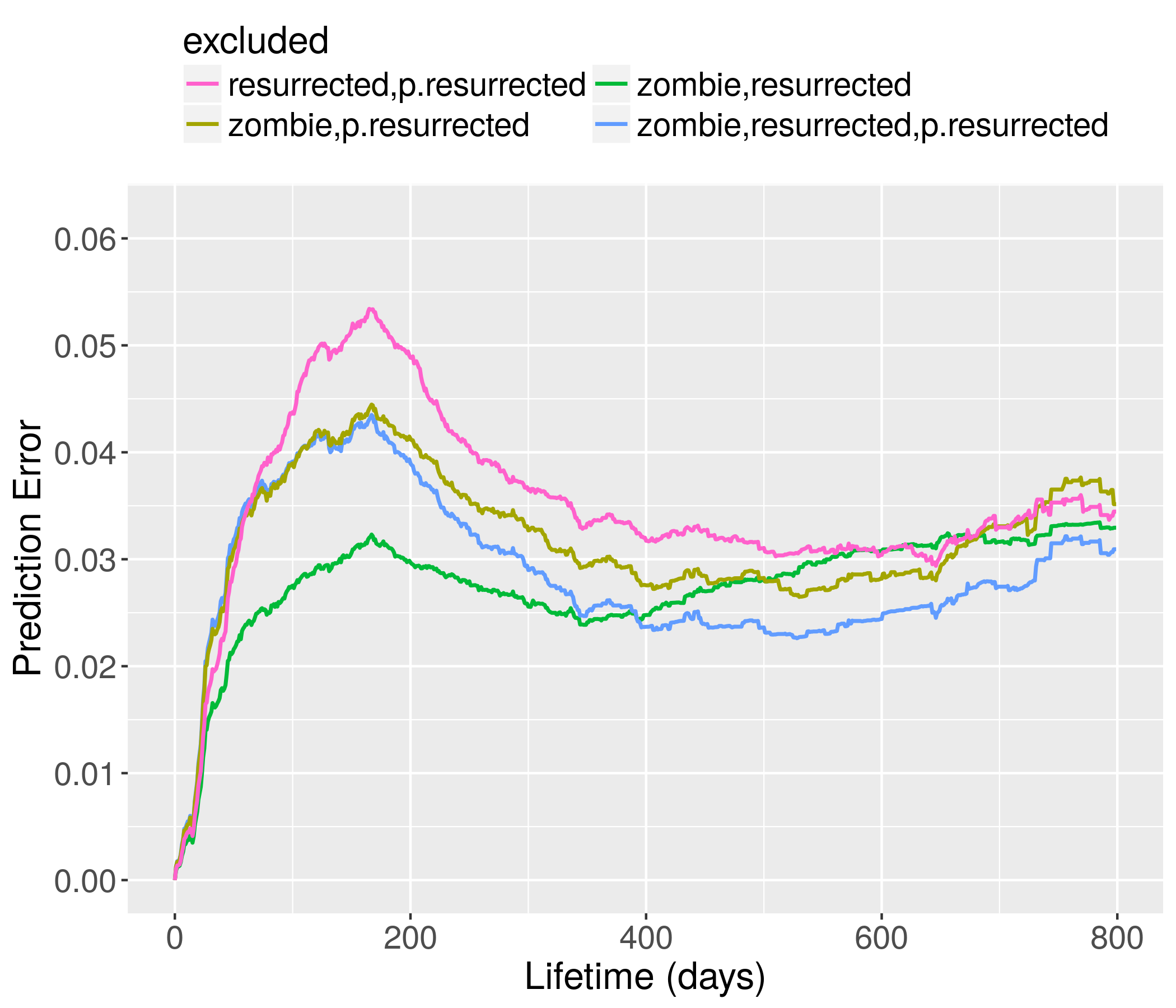}\hfill\null\\
\hfill
\includegraphics[width=0.44\textwidth]{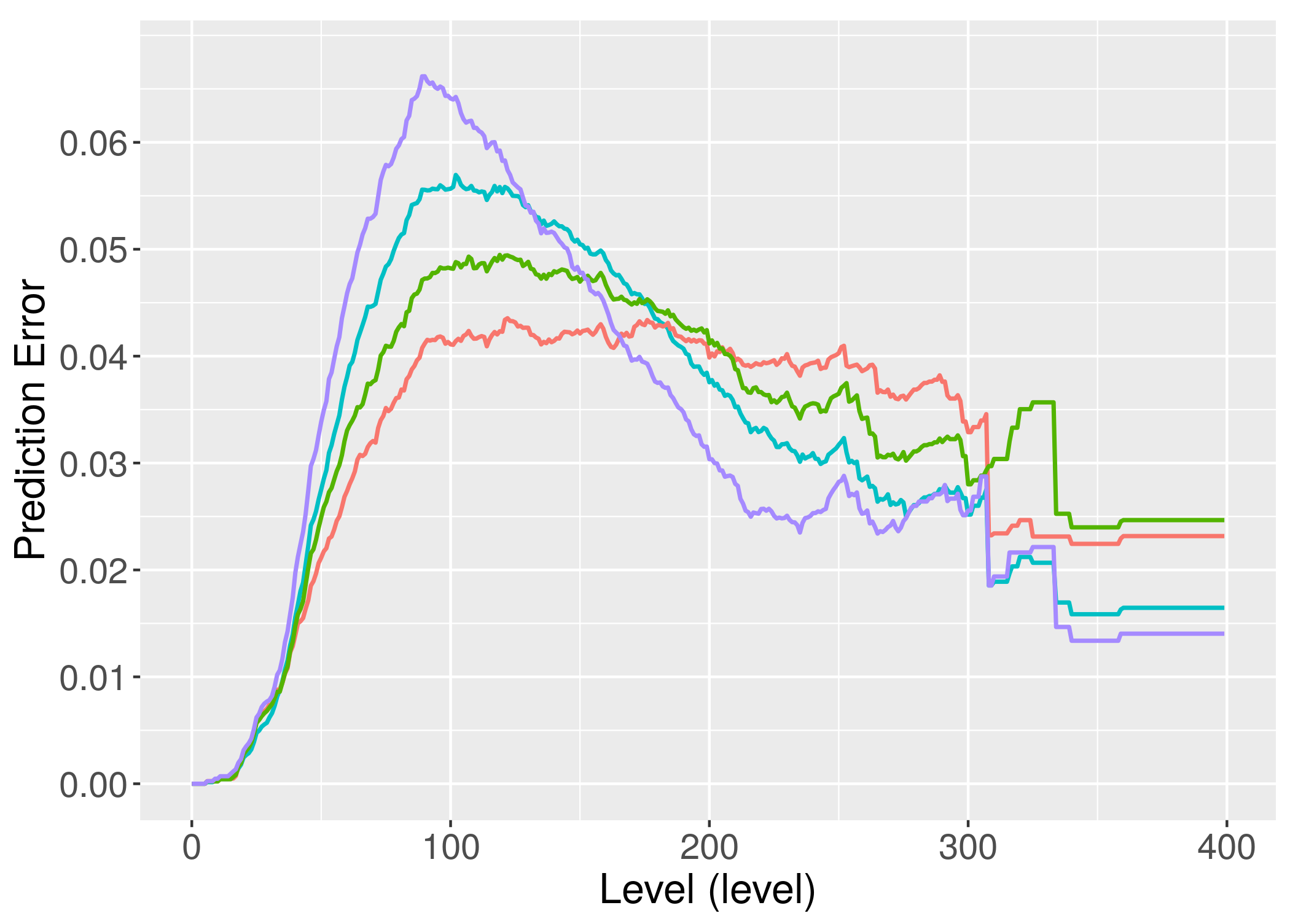}\hfill
\includegraphics[width=0.44\textwidth]{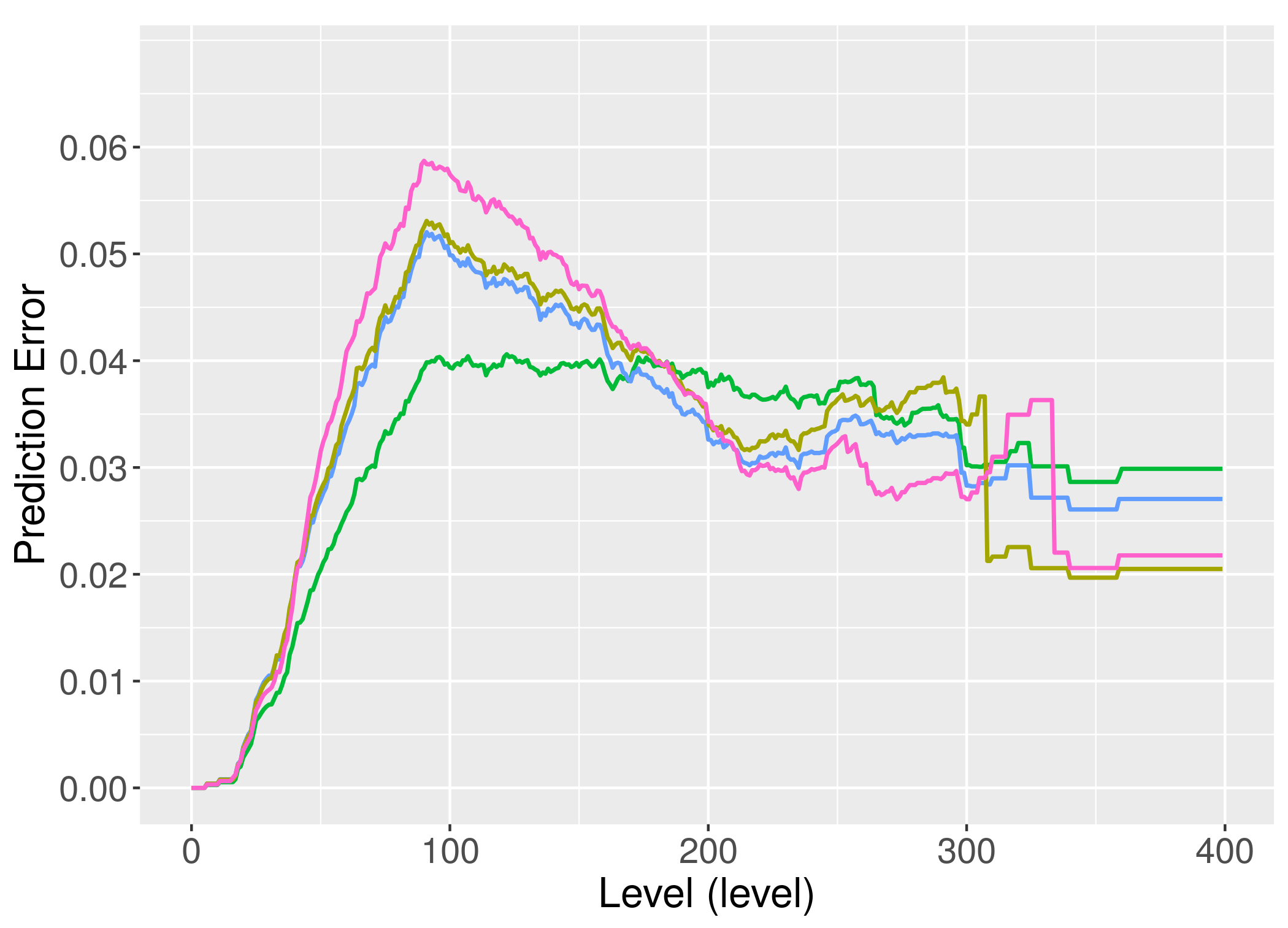}\hfill\null\\
\hfill
\includegraphics[width=0.44\textwidth]{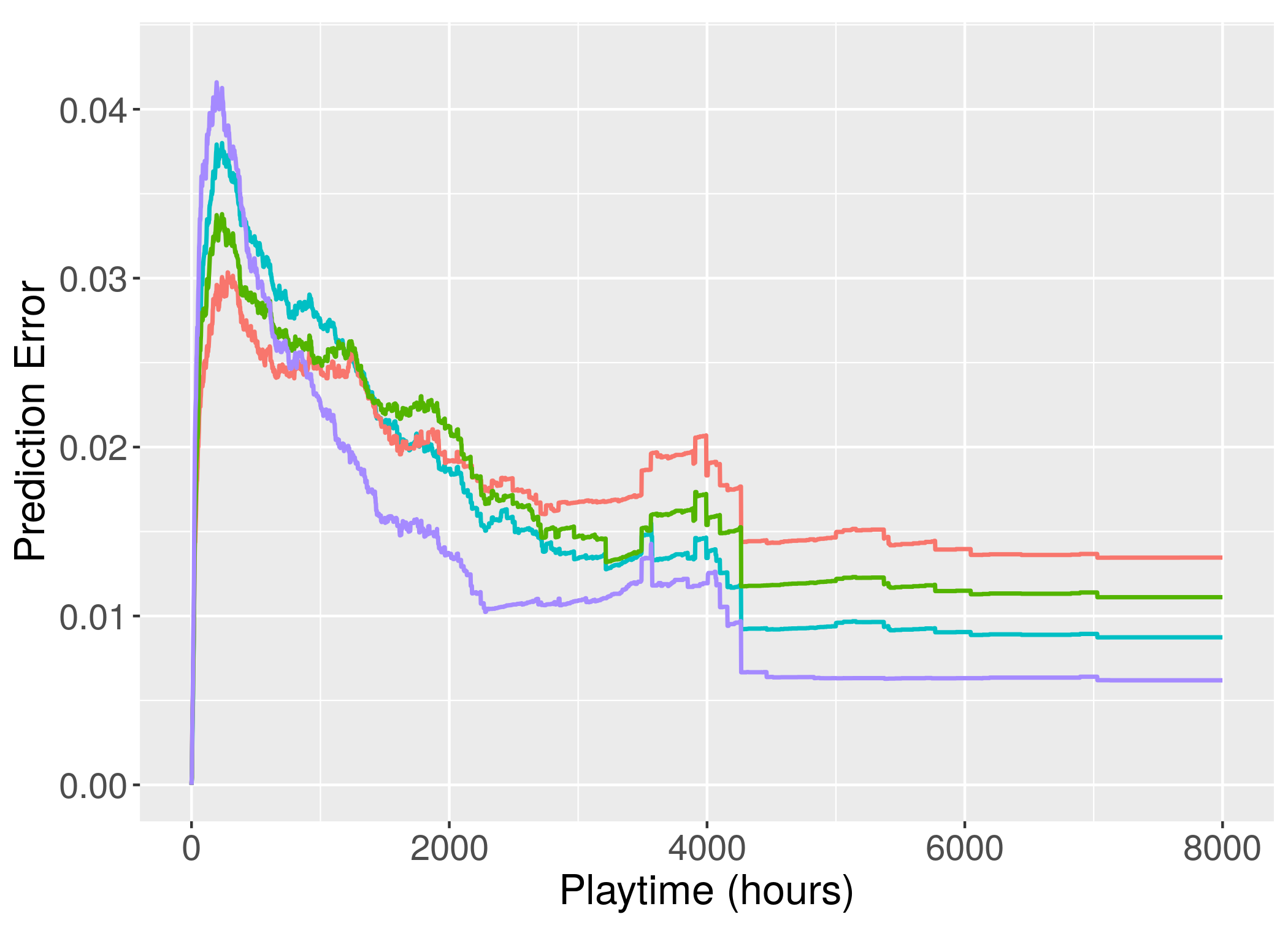}\hfill 
\includegraphics[width=0.44\textwidth]{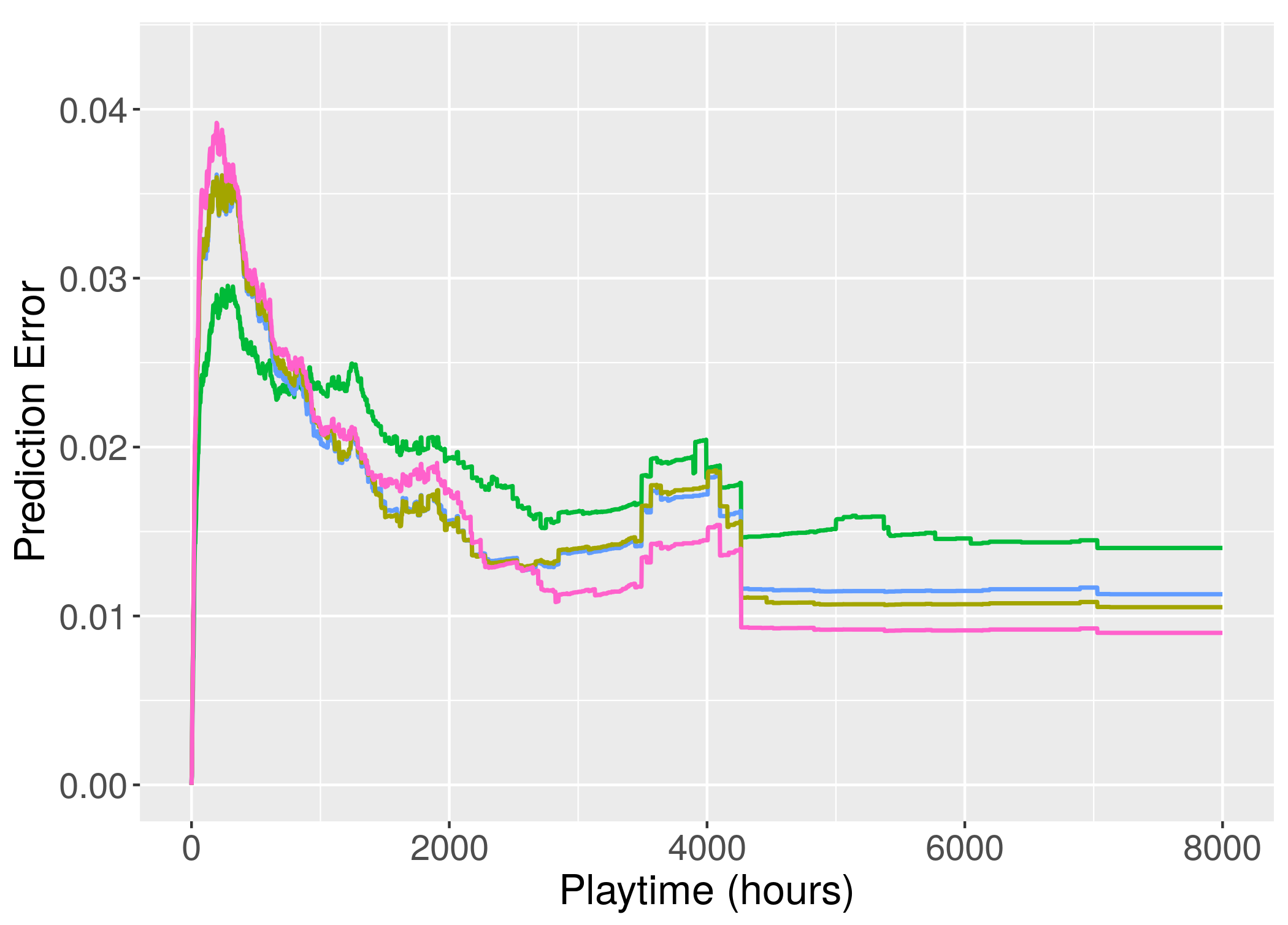}\hfill\null\\
\caption{Prediction error curves for \emph{login} churn as a function of lifetime (top), game level (center) and playtime (bottom). They have been computed using a conditional inference survival ensemble model, upon excluding zombie, resurrected or purchase resurrected players (left) and combinations thereof (right) from the training sample.} 
\label{IBS}
\end{figure*}

\begin{figure*}[ht!]
  \centering
\hfill  
\includegraphics[width=0.44\textwidth]{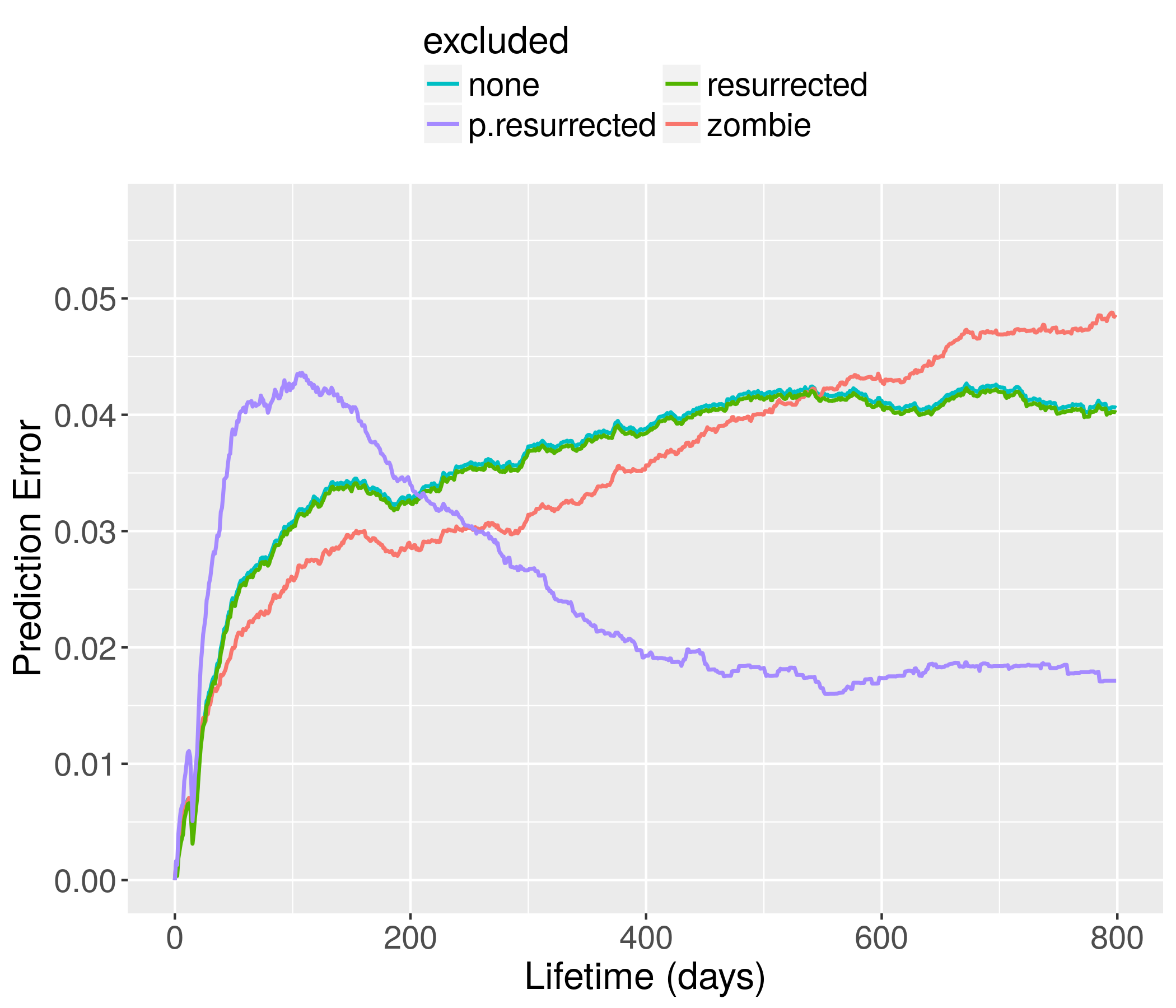}\hfill 
\includegraphics[width=0.44\textwidth]{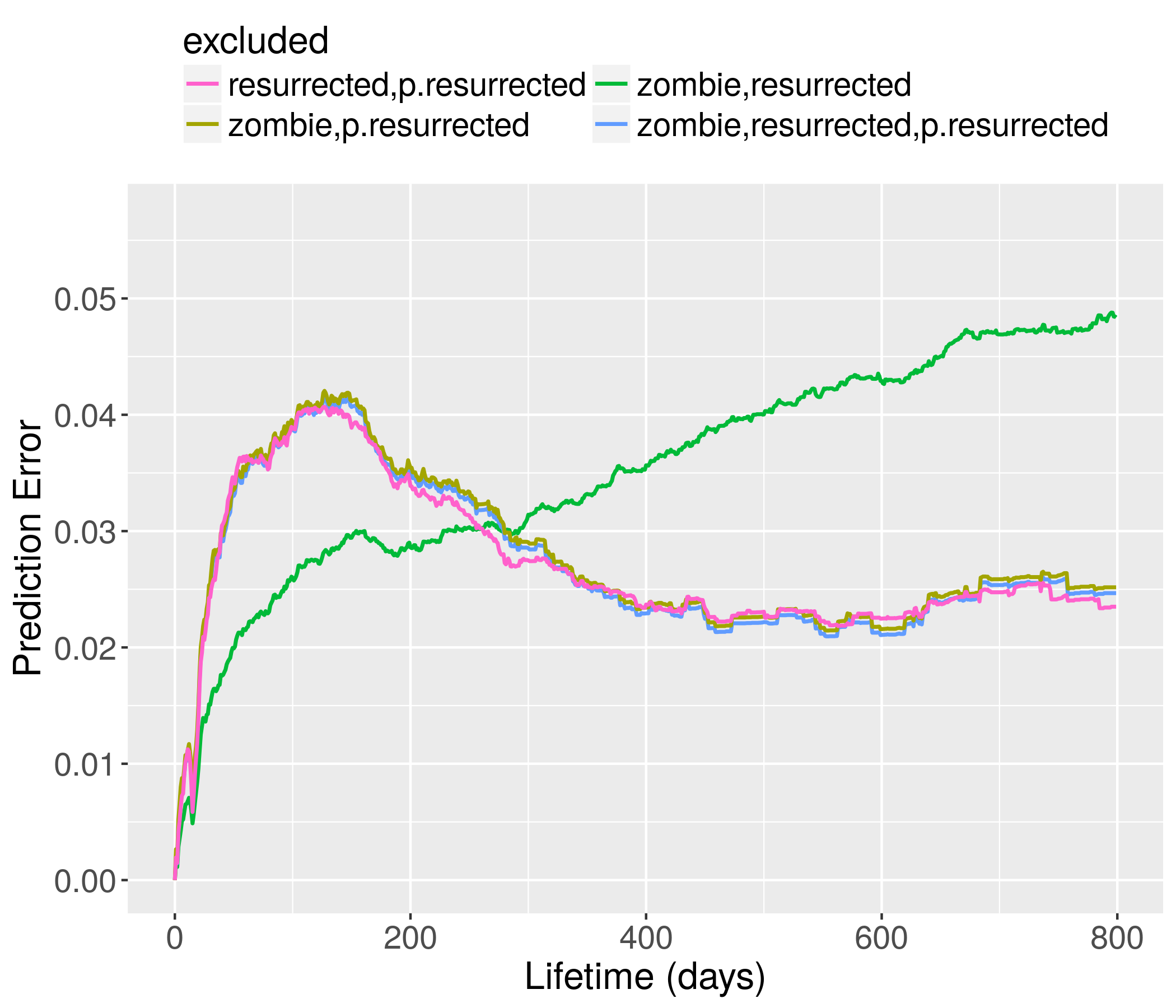}\hfill\null\\
\hfill
\includegraphics[width=0.44\textwidth]{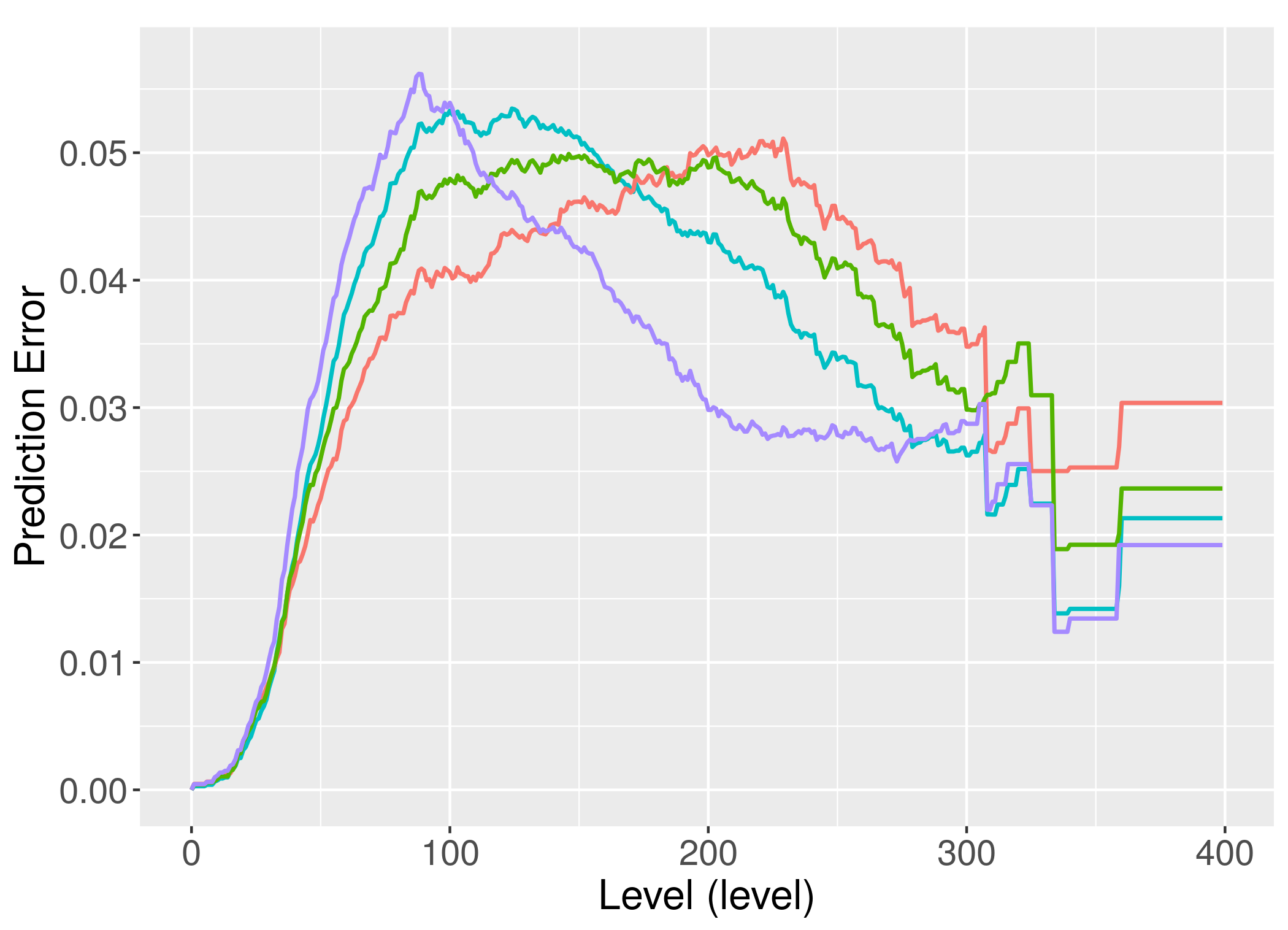}\hfill 
\includegraphics[width=0.44\textwidth]{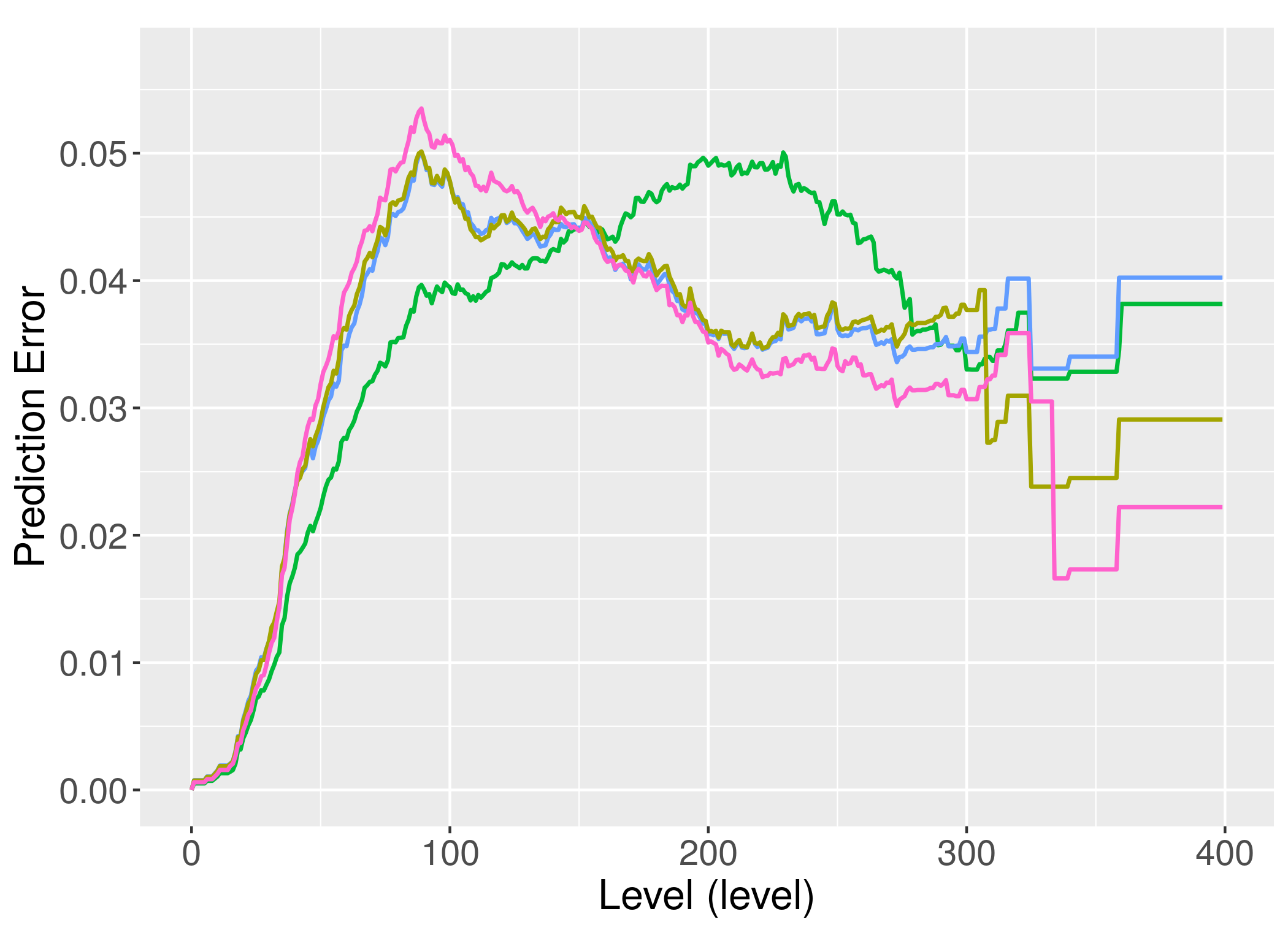}\hfill\null\\
\hfill
\includegraphics[width=0.44\textwidth]{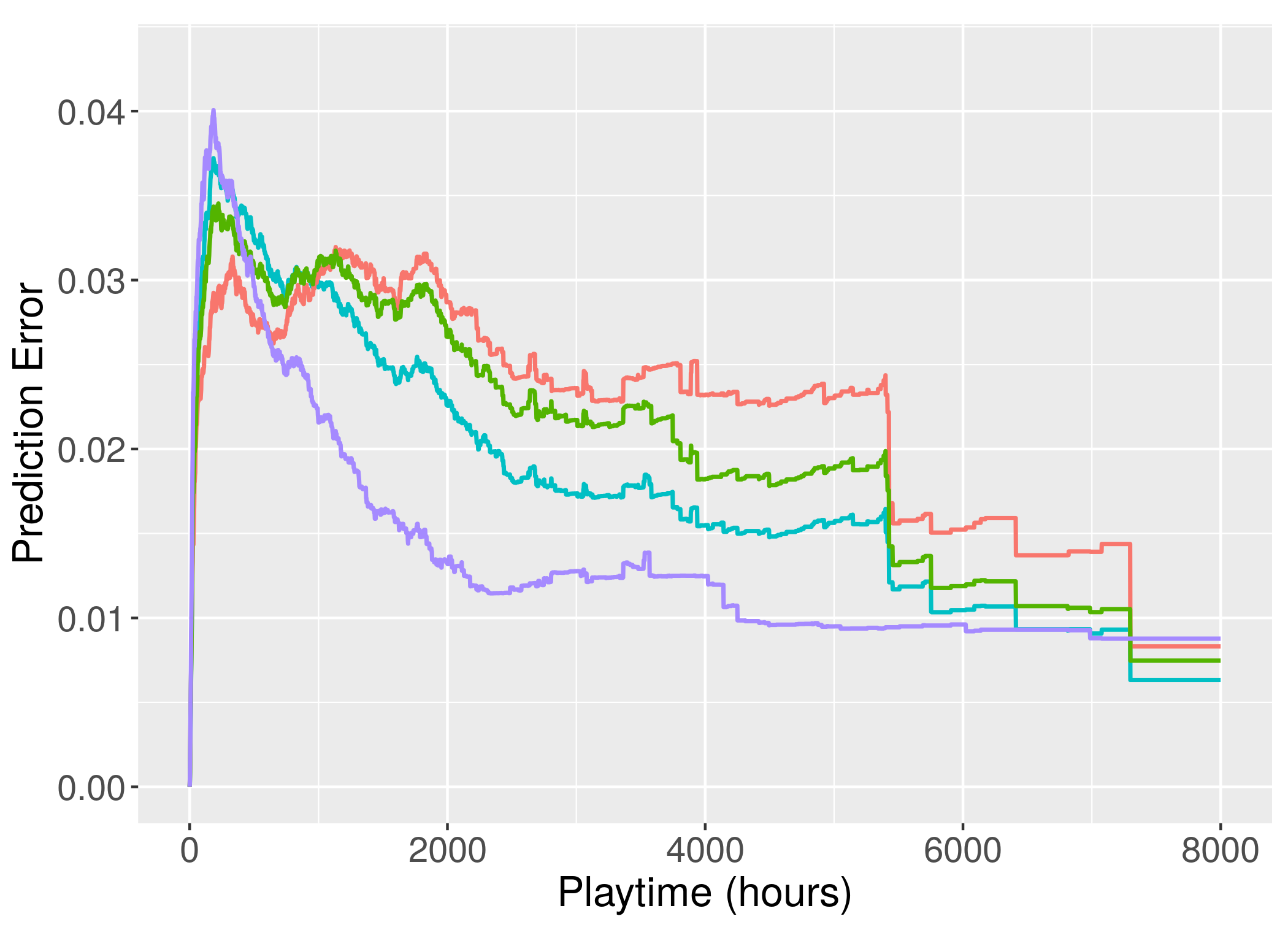}\hfill 
\includegraphics[width=0.44\textwidth]{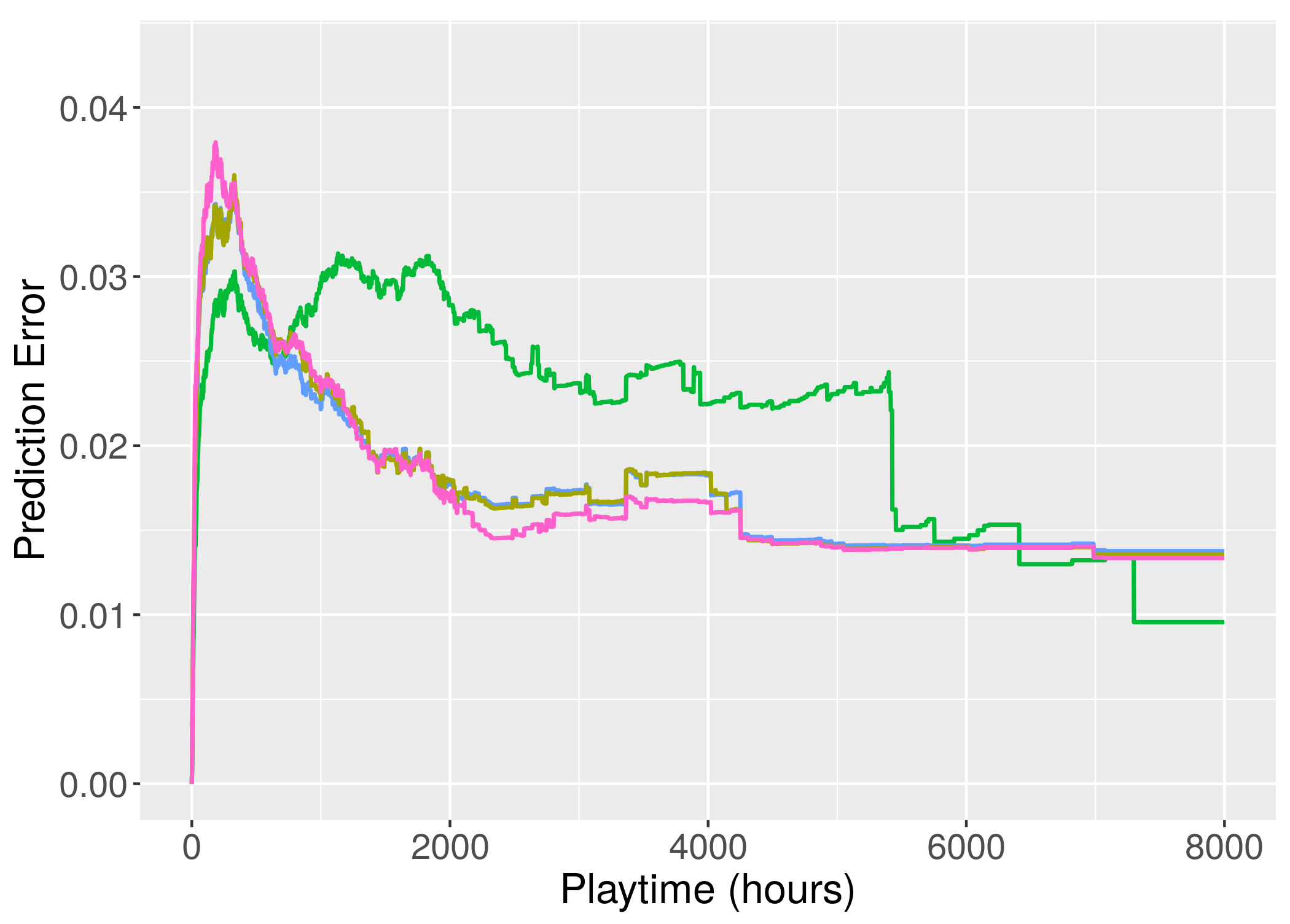}\hfill\null
\caption{Prediction error curves for \emph{purchase} churn as a function of lifetime (top), game level (center) and playtime (bottom). They have been computed using a conditional inference survival ensemble model, upon excluding zombies, resurrected or purchase resurrected players (left) and combinations thereof (right) from the training sample.}
\label{IBS_purchase}
\end{figure*}

The login and purchase churn prediction results for the different models (binary and survival) and survival variables (lifetime, level and playtime) are summarized in Table~\ref{resultsTable}. Prediction error curves from the survival analysis of churn and purchase churn are shown in Figures \ref{IBS} and \ref{IBS_purchase}, respectively. 
Both the table and the figures explore how the prediction accuracy of the models varies when we exclude one or several of the previously described player groups (zombies, resurrected, purchase resurrected) from the training sample. 

The impact of including or excluding these groups is large in the survival analysis, but small to non-existent in the binary classification (where the only action that seems to have a relatively noticeable effect is removing purchase resurrected players when predicting purchase churn). This seems reasonable, as the former method relies on learning probabilities throughout the whole lifetime of each player and is thus much more sensitive to the noise introduced by erratic churn behaviors. 
Remarkably, the choices that optimize the survival results (discussed in detail in what follows) have a negligible to slightly positive impact on the binary models, and thus the same approach could be safely taken for both the classification and survival problems.

Focusing on the left column of Figure~\ref{IBS} (where only individual groups of players have been excluded from the training) we see that, for small lifetime, level and playtime values, the most significant error reduction in login churn predictions is achieved by removing zombies (although there is no such reduction for very short lifetimes), which is also reflected in the IBS scores in Table~\ref{resultsTable}. The improvement is further enhanced as lifetime increases; for high playtime and level, however, the trend is reversed and errors are lower (albeit not significantly) when considering all players. 
Removal of (only) resurrected players exhibits similar patterns, but with a generally lower impact. Curiously, discarding purchase resurrected players has almost the opposite effect: it affects very negatively the performance for small values of all three survival variables, but improves it at large scales---to the point of yielding the best results for high level and playtime. However, the IBS values in Table~\ref{resultsTable} clearly indicate that the overall performance is degraded when removing these users. 

As suggested by the previous discussion, the best overall results for login churn prediction are obtained by excluding both zombies and resurrected (but not purchase resurrected) players from the training sample; see Table~\ref{resultsTable}. On the other hand, the overall negative impact of removing purchase churners can be deemed reasonable, which may be explained by the fact that these players---despite going for long periods without any spending---can maintain typical activity levels in terms of session frequency and duration and in-game progression, thus providing the models with additional valuable information to learn from.

Turning now to purchase churn, the effects of excluding only zombies or only resurrected players (see Figure~\ref{IBS_purchase}, left column) are qualitatively similar to the ones discussed for login churn. 
Purchase resurrected players could have been anticipated to play a major role in understanding purchase churn, 
and indeed their exclusion does provide an overall improvement in all variables (lifetime, level and playtime) as shown by the IBS values in Table~\ref{resultsTable}. 
Interestingly, discarding these players has a negative impact for short lifetimes---an effect compensated for by the great gains at large values.
This could be suggesting that a more restrictive definition of purchase resurrected players 
(by requiring them to start purchasing again after periods not just longer but \emph{much longer} than the purchase churn definition, namely the approach followed for login churn) might be needed. 

As in the case of login churn, excluding both resurrected and zombie players yields good results in terms of lifetime and level; however, for playtime it is better to consider all players. 
The highest overall accuracy is achieved by discarding zombies and purchase resurrected players (being almost irrelevant 
whether or not resurrected players are also discarded). 

\section{SUMMARY AND CONCLUSION}
\label{conclusion}

This study shows that excluding certain types of players (with a particular behavior regarding churn) from the training sample can lead to better churn predictions in the context of video games. Both binary classification and survival models were evaluated. Even though both approaches yield accurate results, the latter seems better suited for churn prediction, since (as discussed in \citealt{perianez2016churn}) it takes into account the censored nature of the problem and provides a much richer output. 
Our results show that, in general, removing active players with very limited activity (zombies) and those who return to the game or after a long period of inactivity (resurrected players) leads to more accurate churn and purchase churn forecasts.
(In the latter case, optimal results are obtained by removing also players who start purchasing again after a long period without spending.) 
Moreover, excluding certain players from the modeling might be helpful from an operational perspective, as it would reduce the size of game datasets.

This work proposes three new types of players based on their churn behavior and aims to establish a basic framework for further related studies (in a similar vein e.g.\ to the already extended use of the ``VIP player'' concept). It also opens new questions in game data science research, such as whether it could be possible to foresee if a certain player will resurrect and how many times she will do so, or to get an accurate time-to-resurrection prediction. Finally, it represents a first step towards finding better and increasingly complex ways to characterize churn behavior that will improve our understanding of the phenomenon and the performance of churn prediction models.

\section*{ACKNOWLEDGEMENT}
\label{acknowledgements}
We thank Javier Grande for reviewing the manuscript.

\bibliography{main}


\end{document}